\documentclass[runningheads,a4paper]{llncs}

\usepackage{graphicx,amsmath,url,amsfonts,booktabs,nicefrac,multirow,microtype,bm,subcaption}

\usepackage{hyperref,cleveref}
\hypersetup{colorlinks=true,linkcolor=blue,urlcolor=blue,citecolor=blue}

\begin{document}

\mainmatter

\title{On the Initialization of Long Short-Term Memory Networks}

\titlerunning{LSTM Initialization}

\authorrunning{Mostafa Mehdipour Ghazi et al.}

\author{Mostafa Mehdipour Ghazi \inst{1,2,3,4} \and Mads Nielsen \inst{1,2,3} \and Akshay Pai \inst{1,2,3} \and Marc Modat \inst{4,5} \and M. Jorge Cardoso \inst{4,5} \and S\'ebastien Ourselin \inst{4,5} \and Lauge S{\o}rensen \inst{1,2,3}}

\institute{Biomediq A/S, Copenhagen, DK, \\
\and Cerebriu A/S, Copenhagen, DK, \\
\and Department of Computer Science, University of Copenhagen, Copenhagen, DK, \\
\and {\fontsize{9}{9}\selectfont Department of Medical Physics and Biomedical Engineering, University College London, London, UK,} \\
\and {\fontsize{9}{9}\selectfont School of Biomedical Engineering and Imaging Sciences, King's College London, London, UK,} \\
\email{mehdipour@biomediq.com}}

\maketitle

\begin{abstract}
Weight initialization is important for faster convergence and stability of deep neural networks training. In this paper, a robust initialization method is developed to address the training instability in long short-term memory (LSTM) networks. It is based on a normalized random initialization of the network weights that aims at preserving the variance of the network input and output in the same range. The method is applied to standard LSTMs for univariate time series regression and to LSTMs robust to missing values for multivariate disease progression modeling. The results show that in all cases, the proposed initialization method outperforms the state-of-the-art initialization techniques in terms of training convergence and generalization performance of the obtained solution.
\end{abstract}

\section{Introduction}

Recurrent neural networks (RNNs) are the state-of-the-art nonparametric methods for sequence learning that map an input sequence to an output sequence by predicting the next time steps. RNN training using the backpropagation through time algorithm is challenging due to vanishing and exploding gradients where the norm of the backpropagated error gradient can increase or decrease exponentially, hindering the network in capturing long-term dependencies \cite{Hochreiter2001}.

Three main solutions have been proposed in the literature to improve RNN training; modifications of the training algorithm, modifications of the network architecture, or different weight initialization schemes. In the first approach, advanced optimization techniques such as the Hessian-Free method \cite{Martens2011} or regularized loss functions \cite{Trinh2018} are applied to improve the backpropagation through time algorithm for learning long sequences. The second approach is to properly initialize the RNN weight matrices, for example, to be identity \cite{Le2015} or orthogonal \cite{Vorontsov2017}, to find solution to the long-term dependency problem. The third approach is to employ nonlinear reset units in the RNN architecture to store information for a long time, for instance, using long short-term memory (LSTM) networks \cite{Hochreiter1997} or gated recurrent units (GRUs) \cite{Cho2014}.

LSTM networks, the most successful type of RNNs, use a gated architecture to replace the hidden unit with a memory cell to efficiently capture long-term temporal dependencies by storing and retrieving sequence information over time. The memory cell is used as a feedback along with three nonlinear (multiplicative) reset units to keep the backpropagated error signal constant. The input and output gates of the cell learn their weights to incorporate the stored information or to control the output values. There is also a forget gate that learns to remember or forget the memory information over time by scaling the cell content. Therefore, in contrast to vanilla RNNs, LSTM units by design allow gradients to flow unchanged, but they can still suffer from instabilities (exploding gradient problem) when trained on long sequences \cite{Sutskever2014}.

In this paper, a simple, yet robust initialization method is proposed to tackle the training instabilities in LSTM networks. The idea is based on normalized random initialization of the network weights with the property that the input and output signals have the same variance. The proposed method is applied to standard LSTMs \cite{Hochreiter2001,Gers2002} for univariate time series regression using data from the UCR Time Series Archive \cite{Dau2018} and to LSTMs robust to missing values \cite{Ghazi2019} for multivariate disease progression modeling in the Alzheimer's Disease Neuroimaging Initiative (ADNI) cohort \cite{Petersen2010} using volumetric magnetic resonance imaging (MRI) measurements.

\section{Related Work}

Since deep neural network training is achieved by solving a nonconvex optimization problem, mostly in a stochastic way, a random weight initialization scheme is important for faster convergence and stability. Otherwise, the magnitudes of the input signal and error gradients at different layers can exponentially decrease or increase, leading to an ill-conditioned problem. Standard initialization of weights with zero-mean uniform/Gaussian distributions and heuristic variances ranging from $0.001$ to $0.01$ or an input layer size ($N$) dependent variance of $1 / (3N)$ have been widely used in previous studies \cite{Glorot2010}. But, studies on the initialization, for instance, using unsupervised pre-training \cite{Erhan2009}, showed its importance as a regularizer for the optimization procedure to robustly reach a local minimum and to improve generalization.

Accordingly, training difficulties have been investigated based on the variance of the responses in each layer, when the singular values of the Jacobian are not unit, and a normalized initialization of uniform weights with a variance of $1 / N$ is suggested assuming that the activation functions are identity and/or hyperbolic tangent \cite{Glorot2010}. Likewise, a scaled initialization method has been developed to train deep rectified models from scratch using zero-mean Gaussian weights whose variances are $2 / N$ \cite{He2015}.

To resolve the long-term temporal dependencies problem in RNNs, which can be seen as deep networks when unfolded through time, the (scaled) identity matrix has been applied to initialize the hidden (recurrent) weights matrix to output the previous hidden state in the absence of the current inputs in RNNs composed of rectified linear units (ReLU) \cite{Le2015}. Alternatively, (nearly) orthogonal matrices \cite{Vorontsov2017} and scaled positive-definite weight matrices \cite{Talathi2015} have been used to address vanishing and exploding gradients in RNNs by preserving the gradient norm during backpropagation.

As it can be seen, different initialization methods have been proposed to deal with the training convergence problem in deep neural networks including RNNs, assuming that LSTMs by design can handle the issue. Hence, the abovementioned initialization methods, e.g., orthogonal recurrent weight matrices and current input weight matrices, both drawn i.i.d. from zero-mean Gaussian distributions with variances of $1 / N$, have also been applied to LSTMs. However, as noted before, LSTMs can still suffer from instability with improper initialization due to the stochastic nature of the optimization and using multiplicative gates and feedback signals.

\section{The Proposed Initialization}

To address training instability and slow convergence in LSTMs, we propose a scaled random weights initialization method that aims to keep the variance of
the network input and output in the same range. Let's $\bm{x}_j^t\in\mathbb{R}^{N\times1}$ be the $j$-th observation of an $N$-dimensional input vector at time $t$. The feedforward pass of an LSTM network can be expressed as
\begin{gather*}
\bm{f}_j^t = \sigma_g(W_f\bm{x}_j^t + U_f\bm{h}_j^{t-1} + \bm{b}_f) \,, \\
\bm{i}_j^t = \sigma_g(W_i\bm{x}_j^t + U_i\bm{h}_j^{t-1} + \bm{b}_i) \,, \\
\bm{z}_j^t = \sigma_c(W_c\bm{x}_j^t + U_c\bm{h}_j^{t-1} + \bm{b}_c) \,, \\
\bm{c}_j^t = \bm{f}_j^t\odot \bm{c}_j^{t-1} + \bm{i}_j^t\odot\bm{z}_j^t \,, \\
\bm{o}_j^t = \sigma_g(W_o\bm{x}_j^t + U_o\bm{h}_j^{t-1} + \bm{b}_o) \,, \\
\bm{h}_j^t = \bm{o}_j^t\odot \sigma_h(\bm{c}_j^t) \,,
\end{gather*}
\noindent where $\{\bm{f}_j^t,\bm{i}_j^t,\bm{z}_j^t,\bm{c}_j^t,\bm{o}_j^t,\bm{h}_j^t\}\in\mathbb{R}^{M\times1}$ are the $j$-th observation of forget gate, input gate, modulation gate, cell state, output gate, and hidden output at time $t$, respectively, and $M$ is the number of output units. Also, $\{W_f,W_i,W_c,W_o\}\in\mathbb{R}^{M\times N}$ are weight matrices containing the connecting weights from input $\bm{x}_j^t$ to the gates and cell, $\{U_f,U_i,U_c,U_o\}\in\mathbb{R}^{M\times M}$ are weight matrices containing the connecting weights from recurrent input $\bm{h}_j^{t-1}$ to the gates and cell, $\{\bm{b}_f,\bm{b}_i,\bm{b}_c,\bm{b}_o\}\in\mathbb{R}^{M\times1}$ denote the corresponding biases of neurons, and $\odot$ is the Hadamard product. Finally, $\sigma_g$, $\sigma_c$, and $\sigma_h$ are nonlinear activation functions allocated to the gates, input modulation, and hidden output, respectively. Note that, in a regression problem, $M = N$, and $\bm{h}_j^{t-1}$ is an estimation of $\bm{x}_j^t$. The regression assumptions can still be applied to sequence-to-sequence or sequence-to-label learning problems simply by adding a fully-connected layer with $N$ input nodes and a desired number of output units.

Assume that all of the weight matrices are independently initialized with zero-mean i.i.d. random values obtained from a symmetric distribution. The goal is to derive the condition(s) on the initialization of the weights to achieve $\mathrm{Var}(\bm{h}_j^{t}) = \mathrm{Var}(\bm{x}_j^t)$. Since the weights are independent from the input, assuming an exact estimation for the recurrent value, i.e., $\bm{h}_j^{t-1} = \bm{x}_j^t$, and mutually independent zero-mean input features -- sharing the same distribution, the variance of the forget gate can be calculated as
\begin{align*}
\begin{split}
\mathrm{Var}(\bm{f}_j^t) & = \mathrm{Var}(\sigma_g(W_f\bm{x}_j^t + U_f\bm{h}_j^{t-1} + \bm{b}_f)) \,, \\
 & = \mathrm{Var}(W_f\bm{x}_j^t + U_f\bm{h}_j^{t-1} + \bm{b}_f) \,, \\
 & = \mathrm{Var}(\left(W_f + U_f\right) \bm{x}_j^{t}) \,, \\
 & = N\left(\mathrm{Var}(w_f) + \mathrm{Var}(u_f)\right) \mathrm{Var}(\bm{x}_j^{t}) \,,
\end{split}
\end{align*}
\noindent where $w_f$ and $u_f$ are the elements of $W_f$ and $U_f$, respectively. The bias in the variance calculation is canceled out as it is an independent constant initialized to zero. Moreover, the second equality holds under the assumption that $\sigma_g$ is an identity function. We will discuss other commonly used functions in LSTM units in Section \ref{active_func}.

Variance calculations for the input, modulation, and output gates can be performed in a similar way to the forget gate. That is to say,
\begin{gather*}
\mathrm{Var}(\bm{i}_j^t) = N\left(\mathrm{Var}(w_i) + \mathrm{Var}(u_i)\right) \mathrm{Var}(\bm{x}_j^{t}) \,, \\
\mathrm{Var}(\bm{z}_j^t) = N\left(\mathrm{Var}(w_c) + \mathrm{Var}(u_c)\right) \mathrm{Var}(\bm{x}_j^{t}) \,, \\
\mathrm{Var}(\bm{o}_j^t) = N\left(\mathrm{Var}(w_o) + \mathrm{Var}(u_o)\right) \mathrm{Var}(\bm{x}_j^{t}) \,,
\end{gather*}
\noindent where $w_i$, $u_i$, $w_c$, $u_c$, $w_o$, and $u_o$ are the elements of $W_i$, $U_i$, $W_c$, $U_c$, $W_o$, and $U_o$, respectively.

The cell state formula is a form of the stochastic recurrence equation \cite{Buraczewski2016}, also known as growing perpetuity, in which the moments of the cell state are time varying. Therefore, one tractable way to stabilize the network training is to set $\mathrm{Var}(\bm{c}_j^t) = \mathrm{Var}(\bm{c}_j^{t-1})$. Accordingly,
\begin{align*}
\begin{split}
\mathrm{Var}(\bm{c}_j^t) & = \mathrm{Var}(\bm{f}_j^t\odot \bm{c}_j^{t-1} + \bm{i}_j^t\odot\bm{z}_j^t) \,, \\
 & = \mathrm{Var}(\bm{f}_j^t) \mathrm{Var}(\bm{c}_j^{t-1}) + \mathrm{Var}(\bm{i}_j^t) \mathrm{Var}(\bm{z}_j^t) \,, \\
 & = \mathrm{Var}(\bm{i}_j^t) \mathrm{Var}(\bm{z}_j^t) / (1-\mathrm{Var}(\bm{f}_j^t)) \,,
\end{split}
\end{align*}
\noindent where the above equation is obtained based on the zero-mean assumption and independence assumption between all of the gates and the cell state to avoid terms containing covariance matrices in the last expression. Also, note that $0 < \mathrm{Var}(\bm{f}_j^t) < 1$.

Finally, the variance of the network output is computed as
\begin{align*}
\begin{split}
\mathrm{Var}(\bm{h}_j^t) & = \mathrm{Var}(\bm{o}_j^t\odot \sigma_h(\bm{c}_j^t)) \,, \\
 & = \mathrm{Var}(\bm{o}_j^t) \mathrm{Var}(\bm{c}_j^t) \,,
\end{split}
\end{align*}
\noindent where the last equality is obtained assuming that there is an identity activation function and independence between the output gate and the cell state. Considering all of the calculated variances and setting $\mathrm{Var}(\bm{h}_j^{t}) = \mathrm{Var}(\bm{x}_j^t) = 1$, the required condition can be summarized as
\begin{equation} \label{myeq1}
\begin{gathered}
0 < \mathrm{Var}(w_f) + \mathrm{Var}(u_f) < 1 / N \,, \\
1 - N \left(\mathrm{Var}(w_f) + \mathrm{Var}(u_f)\right) = \prod_{k \neq f} N \left(\mathrm{Var}(w_k) + \mathrm{Var}(u_k)\right) \,,
\end{gathered}
\end{equation}
\noindent where the right hand side of the above equation is the multiplication of the weights connected to the input, modulation, and output gates.

Similar to the feedforward pass, some initialization conditions can be derived to ensure that the variance of the backpropagated gradient remains unchanged, i.e., $\mathrm{Var}(\partial\mathcal{L} / \partial\bm{h}_j^t) = \mathrm{Var}(\partial\mathcal{L} / \partial\bm{x}_j^t)$ where $\mathcal{L}\in\mathbb{R}$ is the loss function defined based on the actual target and network output. However, as shown in \cite{Glorot2010} and \cite{He2015}, initialization with properly scaling the forward signal is equivalent to initialization with properly scaling the backward signal, and since the number of units in the input and output of the LSTM network are the same, similar conditions for weight initialization using backpropagation will be obtained.

\subsection{Peephole Connections}

In general, LSTMs can be extended to augment their internal cell state to the multiplicative gates using the so-called peephole connections. These cell-to-gate connections allow the gates to inspect the current cell state even if the output gate is closed, and consequently help improving the performance, especially when the task involves a precise duration of intervals \cite{Gers2002}. The feedforward pass of the peephole LSTM can be formulated as
\begin{gather*}
\bm{f}_j^t = \sigma_g(W_f\bm{x}_j^t + U_f\bm{h}_j^{t-1} + V_f\bm{c}_j^{t-1} + \bm{b}_f) \,, \\
\bm{i}_j^t = \sigma_g(W_i\bm{x}_j^t + U_i\bm{h}_j^{t-1} + V_i\bm{c}_j^{t-1} + \ \bm{b}_i) \,, \\
\bm{z}_j^t = \sigma_c(W_c\bm{x}_j^t + U_c\bm{h}_j^{t-1} + \bm{b}_c) \,, \\
\bm{c}_j^t = \bm{f}_j^t\odot \bm{c}_j^{t-1} + \bm{i}_j^t\odot\bm{z}_j^t \,, \\
\bm{o}_j^t = \sigma_g(W_o\bm{x}_j^t + U_o\bm{h}_j^{t-1} + V_o\bm{c}_j^{t} + \ \bm{b}_o) \,, \\
\bm{h}_j^t = \bm{o}_j^t\odot \sigma_h(\bm{c}_j^t) \,,
\end{gather*}

\noindent where $\{V_f,V_i,V_o\}\in\mathbb{R}^{M\times M}$ are diagonal peephole weight matrices. Hence, each gate will only look at its corresponding cell state. To achieve $\mathrm{Var}(\bm{h}_j^{t}) = \mathrm{Var}(\bm{x}_j^t)$, all the assumptions applied to the traditional LSTM are used for the peephole LSTM. Assuming that the peephole matrices are independent from the input and the cell state and are independently initialized with zero-mean i.i.d. random values obtained from a symmetric distribution, the variances can be calculated as
\begin{gather}
\mathrm{Var}(\bm{f}_j^t) = N\left(\mathrm{Var}(w_f) + \mathrm{Var}(u_f)\right) \mathrm{Var}(\bm{x}_j^{t}) + \mathrm{Var}(v_f) \mathrm{Var}(\bm{c}_j^{t-1}) \,, \label{eq1} \\
\mathrm{Var}(\bm{i}_j^t) = N\left(\mathrm{Var}(w_i) + \mathrm{Var}(u_i)\right) \mathrm{Var}(\bm{x}_j^{t}) + \mathrm{Var}(v_i) \mathrm{Var}(\bm{c}_j^{t-1}) \,, \label{eq2} \\
\mathrm{Var}(\bm{z}_j^t) = N\left(\mathrm{Var}(w_c) + \mathrm{Var}(u_c)\right) \mathrm{Var}(\bm{x}_j^{t}) \,, \label{eq3} \\
\mathrm{Var}(\bm{o}_j^t) = N\left(\mathrm{Var}(w_o) + \mathrm{Var}(u_o)\right) \mathrm{Var}(\bm{x}_j^{t}) + \mathrm{Var}(v_o) \mathrm{Var}(\bm{c}_j^{t}) \,, \label{eq4} \\
\mathrm{Var}(\bm{c}_j^t) = \mathrm{Var}(\bm{c}_j^{t-1}) = \mathrm{Var}(\bm{i}_j^t) \mathrm{Var}(\bm{z}_j^t) / (1-\mathrm{Var}(\bm{f}_j^t)) \,, \label{eq5} \\
\mathrm{Var}(\bm{h}_j^t) = \mathrm{Var}(\bm{o}_j^t) \mathrm{Var}(\bm{c}_j^t) \,, \label{eq6}
\end{gather}
\noindent where $v_f$, $v_i$, and $v_o$ are the diagonal elements of $V_f$, $V_i$, and $V_o$, respectively. Merging Equations (\ref{eq4}) and (\ref{eq6}) under the assumption that $\mathrm{Var}(\bm{h}_j^{t}) = \mathrm{Var}(\bm{x}_j^t) = 1$ results in a quadratic equation that can be expressed as
\begin{gather} \label{quad1}
\beta_{01} + \beta_{11} \mathrm{Var}(\bm{c}_j^t) + \beta_{21} \mathrm{Var}^2(\bm{c}_j^t) = 0 \,,
\end{gather}
\noindent where $\beta_{01} = -1$, $\beta_{11} = N \left(\mathrm{Var}(w_o) + \mathrm{Var}(u_o)\right)$, and $\beta_{21} = \mathrm{Var}(v_o)$. Since the discriminant $\Delta_1 = \beta_{11}^2 - 4 \beta_{21} \beta_{01}$ is always positive considering nonzero variances, there are two possible solutions for Equation (\ref{quad1}): $\mathrm{Var}(\bm{c}_j^t) = (-\beta_{11} \pm \sqrt{\Delta_1}) / (2\beta_{21})$. However, since $\beta_{21} > 0$ and $\beta_{01} < 0$, with a positive discriminant and based on the sign of the product of the roots ($\beta_{01} / \beta_{21}$), one of the real solutions would be negative, which cannot be accepted as $\mathrm{Var}(\bm{c}_j^t)  > 0$. Therefore, the desired solution to Equation (\ref{quad1}) will be obtained as
\begin{gather} \label{sol1}
\mathrm{Var}(\bm{c}_j^t) = \frac{-\beta_{11} + \sqrt{\Delta_1}}{2\beta_{21}} \,.
\end{gather}

Likewise, combining \Crefrange{eq1}{eq3} and (\ref{eq5}) using the same assumptions leads to another quadratic equation that can be written as
\begin{gather} \label{quad2}
\beta_{02} + \beta_{12} \mathrm{Var}(\bm{c}_j^t) + \beta_{22} \mathrm{Var}^2(\bm{c}_j^t) = 0 \,,
\end{gather}
\noindent where $\beta_{02} = N^2 \left(\mathrm{Var}(w_i) + \mathrm{Var}(u_i)\right) \left(\mathrm{Var}(w_c) + \mathrm{Var}(u_c)\right)$, $\beta_{22} = \mathrm{Var}(v_f)$, and $\beta_{12} = N \mathrm{Var}(v_i) \left(\mathrm{Var}(w_c) + \mathrm{Var}(u_c)\right) + N \left(\mathrm{Var}(w_f) + \mathrm{Var}(u_f)\right) - 1$. The two possible solutions for Equation (\ref{quad2}) will be obtained as $\mathrm{Var}(\bm{c}_j^t) = (-\beta_{12} \pm \sqrt{\Delta_2}) / (2\beta_{22})$, where $\Delta_2 = \beta_{12}^2 - 4 \beta_{22} \beta_{02}$ is the discriminant of the equation. Here, since $\beta_{02},\beta_{22} > 0$, assuming a nonnegative discriminant and based on the sign of the sum and product of the roots ($-\beta_{12} / \beta_{22}$ and $\beta_{02} / \beta_{22}$), both real solutions could be positive and acceptable provided that $\beta_{12} < 0$. However, to achieve a simple solution for initialization, one can set $\Delta_2 = 0$ and $\beta_{12} < 0$ which produces repeated real positive roots for the problem. Therefore, the real solution to Equation (\ref{quad2}) can be obtained as
\begin{gather} \label{sol2}
\mathrm{Var}(\bm{c}_j^t) = \frac{-\beta_{12}}{2\beta_{22}} \,.
\end{gather}

Finally, conditions for the existence of a common solution to Equations (\ref{quad1}) and (\ref{quad2}) can be obtained using Equations (\ref{sol1}) and (\ref{sol2}) as follows
\begin{equation} \label{myeq2}
\begin{gathered}
0 < \mathrm{Var}(v_i) \left(\mathrm{Var}(w_c) + \mathrm{Var}(u_c)\right) + \left(\mathrm{Var}(w_f) + \mathrm{Var}(u_f)\right) < 1/N \,, \\
\frac{\mathrm{Var}(v_o)}{\mathrm{Var}(v_f)} \sqrt{4 N^2 \mathrm{Var}(v_f) \left(\mathrm{Var}(w_i) + \mathrm{Var}(u_i)\right) \left(\mathrm{Var}(w_c) + \mathrm{Var}(u_c)\right)} = \\
\sqrt{N^2 \left(\mathrm{Var}(w_o) + \mathrm{Var}(u_o)\right)^2 + 4 \mathrm{Var}(v_o)} - N \left(\mathrm{Var}(w_o) + \mathrm{Var}(u_o)\right) \,.
\end{gathered}
\end{equation}

\subsection{Nonlinear Activation Functions} \label{active_func}

All the abovementioned equations are obtained based on the assumption that the activation functions are identity functions. In general, symmetric functions with zero intercepts such as the identity and hyperbolic tangent are suggested for $\sigma_h$ and $\sigma_c$, respectively, and logistic sigmoid is suggested for $\sigma_g$ \cite{Gers2002}. Both the hyperbolic tangent and logistic sigmoid are nonlinear symmetric functions that can be linearly approximated using a Taylor series expansion. The former has a zero intercept and its expansion about zero leads to an identity function ($\sigma_c(x) \approx x$). The latter, however, has a nonzero intercept and its Taylor series about zero is approximated as $\sigma_g(x) \approx 0.5 + 0.25x$. Therefore, the sigmoid function approximately increases the input signal mean by $1/2$ and scales its variance by $1/16$. Note that the nonzero mean value of the sigmoid can induce important singular values in the Hessian matrix, resulting in saturation of the top layers and prohibition of gradients to flow backward to learn useful features in the lower layers \cite{Glorot2010}. Using the suggested activation functions in the gates, the variance calculations for the traditional LSTM network are updated as follows based on the aforementioned Taylor series expansion
\begin{gather*}
\mathrm{Var}(\bm{f}_j^t) = N\left(\mathrm{Var}(w_f) + \mathrm{Var}(u_f)\right) \mathrm{Var}(\bm{x}_j^{t})/16 \,, \\
\mathrm{Var}(\bm{i}_j^t) = N\left(\mathrm{Var}(w_i) + \mathrm{Var}(u_i)\right) \mathrm{Var}(\bm{x}_j^{t})/16 \,, \\
\mathrm{Var}(\bm{z}_j^t) = N\left(\mathrm{Var}(w_c) + \mathrm{Var}(u_c)\right) \mathrm{Var}(\bm{x}_j^{t}) \,, \\
\mathrm{Var}(\bm{o}_j^t) = N\left(\mathrm{Var}(w_o) + \mathrm{Var}(u_o)\right) \mathrm{Var}(\bm{x}_j^{t})/16 \,, \\
\mathrm{Var}(\bm{c}_j^t) = \mathrm{Var}(\bm{c}_j^{t-1}) = (\mathrm{Var}(\bm{i}_j^t)+0.25) \mathrm{Var}(\bm{z}_j^t) / (0.75-\mathrm{Var}(\bm{f}_j^t)) \,,
\end{gather*}
\noindent where the last equation is obtained bearing in mind that $\mathrm{Var}(xy) = \mathrm{Var}(x) \mathrm{Var}(y) + \mathbb{E}^2(x) \mathrm{Var}(y) + \mathbb{E}^2(y) \mathrm{Var}(x)$ for two independent random variables $x$ and $y$, and considering $\mathbb{E}(\bm{z}_j^t) = 0$, $\mathbb{E}(\bm{f}_j^t) = \mathbb{E}(\bm{i}_j^t) = 0.5$, and, hence, $\mathbb{E}(\bm{c}_j^t) = \mathbb{E}(\bm{c}_j^{t-1}) = 0$. Finally, the updated rule for initialization of a traditional LSTM network using Equation (\ref{eq6}) can be written as
\begin{equation} \label{myeq3}
\resizebox{0.91\hsize}{!}{$\begin{gathered}
0 < \mathrm{Var}(w_f) + \mathrm{Var}(u_f) < 12 / N \,, \\
\frac{12 - N \left(\mathrm{Var}(w_f) + \mathrm{Var}(u_f)\right)}{N \left(\mathrm{Var}(w_i) + \mathrm{Var}(u_i)\right) + 4} = N^2 \left(\mathrm{Var}(w_o) + \mathrm{Var}(u_o)\right) \left(\mathrm{Var}(w_c) + \mathrm{Var}(u_c)\right) / 16 \,.
\end{gathered}$}
\end{equation}

Applying the same suggested functions in the peephole LSTM network generalizes the variance calculations as follows
\begin{gather*}
\mathrm{Var}(\bm{f}_j^t) = N\left(\mathrm{Var}(w_f) + \mathrm{Var}(u_f)\right) \mathrm{Var}(\bm{x}_j^{t})/16 + \mathrm{Var}(v_f) \mathrm{Var}(\bm{c}_j^{t-1})/16 \,, \\
\mathrm{Var}(\bm{i}_j^t) = N\left(\mathrm{Var}(w_i) + \mathrm{Var}(u_i)\right) \mathrm{Var}(\bm{x}_j^{t})/16 + \mathrm{Var}(v_i) \mathrm{Var}(\bm{c}_j^{t-1})/16 \,, \\
\mathrm{Var}(\bm{z}_j^t) = N\left(\mathrm{Var}(w_c) + \mathrm{Var}(u_c)\right) \mathrm{Var}(\bm{x}_j^{t}) \,, \\
\mathrm{Var}(\bm{o}_j^t) = N\left(\mathrm{Var}(w_o) + \mathrm{Var}(u_o)\right) \mathrm{Var}(\bm{x}_j^{t})/16 + \mathrm{Var}(v_o) \mathrm{Var}(\bm{c}_j^{t})/16 \,, \\
\mathrm{Var}(\bm{c}_j^t) = \mathrm{Var}(\bm{c}_j^{t-1}) = (\mathrm{Var}(\bm{i}_j^t)+0.25) \mathrm{Var}(\bm{z}_j^t) / (0.75-\mathrm{Var}(\bm{f}_j^t)) \,,
\end{gather*}

Here also using Equation (\ref{eq6}), two quadratic equations can be obtained similar to Equations (\ref{quad1}) and (\ref{quad2}), where $\beta_{01} = -16$, $\beta_{11} = N \left(\mathrm{Var}(w_o) + \mathrm{Var}(u_o)\right)$, $\beta_{21} = \mathrm{Var}(v_o)$, $\beta_{02} = N \left(\mathrm{Var}(w_c) + \mathrm{Var}(u_c)\right) (N \left(\mathrm{Var}(w_i) + \mathrm{Var}(u_i)\right) + 4)$, $\beta_{22} = \mathrm{Var}(v_f)$, and $\beta_{12} = N \mathrm{Var}(v_i) \left(\mathrm{Var}(w_c) + \mathrm{Var}(u_c)\right) + N \left(\mathrm{Var}(w_f) + \mathrm{Var}(u_f)\right) - 12$. Likewise, conditions for the existence of a common solution to Equations (\ref{quad1}) and (\ref{quad2}) can be obtained using Equations (\ref{sol1}) and (\ref{sol2}) as follows
\begin{equation} \label{myeq4}
\begin{gathered}
0 < \mathrm{Var}(v_i) \left(\mathrm{Var}(w_c) + \mathrm{Var}(u_c)\right) + \left(\mathrm{Var}(w_f) + \mathrm{Var}(u_f)\right) < 12/N \,, \\
\frac{\mathrm{Var}(v_o)}{\mathrm{Var}(v_f)} \sqrt{4 N \mathrm{Var}(v_f) \left(\mathrm{Var}(w_c) + \mathrm{Var}(u_c)\right) (N \left(\mathrm{Var}(w_i) + \mathrm{Var}(u_i)\right) + 4)} = \\
\sqrt{N^2 \left(\mathrm{Var}(w_o) + \mathrm{Var}(u_o)\right)^2 + 64 \mathrm{Var}(v_o)} - N \left(\mathrm{Var}(w_o) + \mathrm{Var}(u_o)\right) \,.
\end{gathered}
\end{equation}

\subsection{Initialization Summary}
 
The proposed initialization rule can be summarized as follows:
\begin{itemize}
\item Standardize the input data to have a zero mean and unit variance per feature, and initialize the LSTM network biases to zero. 
\item Initialize the weights in the weight matrices randomly using zero-mean i.i.d. Gaussian distributions with variances satisfying one of the following equations:
\begin{itemize}
\item Equation (\ref{myeq1}), if using the traditional LSTM network based on identity or hyperbolic tangent functions.
\item Equation (\ref{myeq2}), if using the peephole LSTM network based on identity or hyperbolic tangent functions.
\item Equation (\ref{myeq3}), if using the traditional LSTM network based on identity or hyperbolic tangent for input modulation and cell activation, and logistic sigmoid functions in the gates.
\item Equation (\ref{myeq4}), if using the peephole LSTM network based on identity or hyperbolic tangent for input modulation and cell activation, and logistic sigmoid functions in the gates.
\end{itemize}
\end{itemize}
Note that the variances need to be selected subject to the specified conditions in the selected equation. For example, when using a peephole LSTM, and, correspondingly, Equation (\ref{myeq2}) or (\ref{myeq4}), there are eleven variances to fix, $\mathrm{Var}(v_f)$, $\mathrm{Var}(v_i)$, $\mathrm{Var}(v_o)$, $\mathrm{Var}(w_f)$, $\mathrm{Var}(u_f)$, $\mathrm{Var}(w_i)$, $\mathrm{Var}(u_i)$, $\mathrm{Var}(w_o)$, $\mathrm{Var}(u_o)$, $\mathrm{Var}(w_c)$, and $\mathrm{Var}(u_c)$.

\section{Experiments and Results}

\subsection{Data}

Both univariate and multivariate data are used to study the effect of initialization on LSTM training.

The following three univariate datasets are obtained from the UCR Time Series Archive \cite{Dau2018} due to having the largest training samples size: \textit{ElectricDevices} with 16,637 samples (8,926 for training and 7,711 for test) of sequence length 96; \textit{FordA} with 4,921 samples (3,601 for training and 1,320 for test) of sequence length 500; and \textit{Crop} with 24,000 samples (7,200 for training and 16,800 for test) of sequence length 46.

The multivariate dataset, \textit{ADNI}, focuses on disease progression modeling and is obtained from the ADNI cohort \cite{Petersen2010}. It constitutes yearly measurements for 383 subjects (332 for training and 51 for test) of sequence length 3 to 10 with normal cognition, mild cognition impairment, or Alzheimer's disease. The multivariate feature set consists of T1-weighted MRI volumetric measurements of ventricles, hippocampus, whole brain, fusiform, middle temporal gyrus, and entorhinal cortex, all normalized for intracranial volume.

\subsection{Experimental Setup}

The proposed initialization method is assessed using a peephole LSTM \cite{Gers2002} applied to the univariate data ($N = 1$) for time series regression and a peephole LSTM robust to missing values \cite{Ghazi2019} applied to the multivariate data ($N = 6$) for disease progression modeling. In both cases, an identity function and a hyperbolic tangent are used in $\sigma_h$ and $\sigma_c$, respectively, a logistic sigmoid is used in $\sigma_g$, and the network biases are initialized to zero. Therefore, the variance selection for weight matrices is performed using Equation (\ref{myeq4}), and weight values are drawn from the zero-mean i.i.d. Gaussian distributions. Four different configurations of the variances are inspected as illustrated in Table~\ref{table0}.

\begin{table}[b]
\caption{The utilized configurations of the variances satisfying Equation (\ref{myeq4}).}
\label{table0}
\centering
\renewcommand{\arraystretch}{1.25}
\tiny
\begin{tabular}{lccccccccccc}
\toprule
Method & $\mathrm{Var}(v_f)$ & $\mathrm{Var}(v_i)$ & $\mathrm{Var}(v_o)$ & $\mathrm{Var}(w_f)$ & $\mathrm{Var}(u_f)$ & $\mathrm{Var}(w_i)$ & $\mathrm{Var}(u_i)$ & $\mathrm{Var}(w_o)$ & $\mathrm{Var}(u_o)$ & $\mathrm{Var}(w_c)$ & $\mathrm{Var}(u_c)$ \\ \hline
\midrule
Proposed 1 & $1$ & $1$ & $1$ & $1/N$ & $1/N$ & $2/N$ & $2/N$ & $3/N$ & $3/N$ & $1/(4N)$ & $1/(4N)$ \\
Proposed 2 & $1/2$ & $1/2$ & $1/2$ & $1/N$ & $1/N$ & $2/N$ & $2/N$ & $1/N$ & $1/N$ & $1/(2N)$ & $1/(2N)$ \\
Proposed 3 & $1$ & $1$ & $1$ & $3/(4N)$ & $1/(4N)$ & $3/N$ & $1/N$ & $4/N$ & $2/N$ & $1/(4N)$ & $1/(4N)$ \\
Proposed 4 & $1$ & $1$ & $1$ & $1/(4N)$ & $3/(4N)$ & $1/N$ & $3/N$ & $2/N$ & $4/N$ & $1/(4N)$ & $1/(4N)$ \\
\bottomrule
\end{tabular}
\end{table}

The input data is standardized to have a zero mean and unit variance per feature dimension. Moreover, the batch size is set to $85\%$ of training samples ($15\%$ used for validation to tune the optimization hyperparameters), and the first to penultimate time point is used to estimate the second to last time point per observation. The L2-norm is used as loss function and momentum batch gradient descent is applied to optimize the network parameters using L2 regularization. The optimization hyperparameters, i.e., the learning rate, momentum weight, and weight decay are set to $0.1$, $0.9$, and $0.0001$, respectively. These values were selected according to the validation set error across the different experiments.

The proposed approach is compared with two state-of-the-art initialization techniques applied to the same LSTM networks assuming zero biases and using the same optimization setup: \textit{normalized} \cite{Glorot2010}, all weight matrices drawn i.i.d. from zero-mean Gaussian distributions with a scaled variance of $1 / N$; \textit{orthogonal} \cite{Vorontsov2017}, same as normalized, but with orthogonal recurrent weight matrices drawn i.i.d. from zero-mean Gaussian distributions with a variance of $1 / N$.

\subsection{Results}

\begin{figure}[t]
\centering
\begin{subfigure}[t]{0.47\textwidth}
\raisebox{-\height}{\includegraphics[scale=0.42]{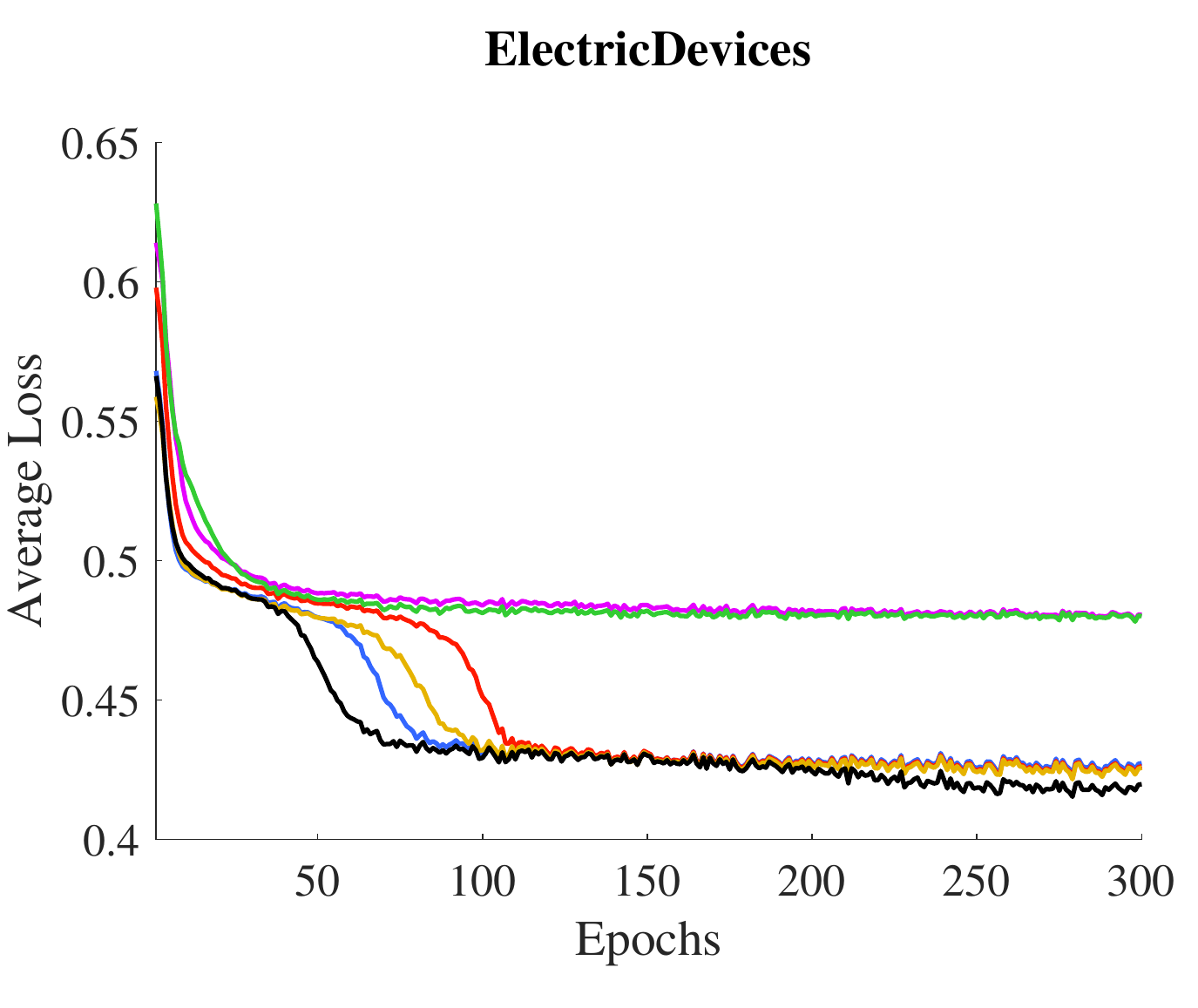}}
\end{subfigure}
\begin{subfigure}[t]{0.47\textwidth}
\raisebox{-\height}{\includegraphics[scale=0.42]{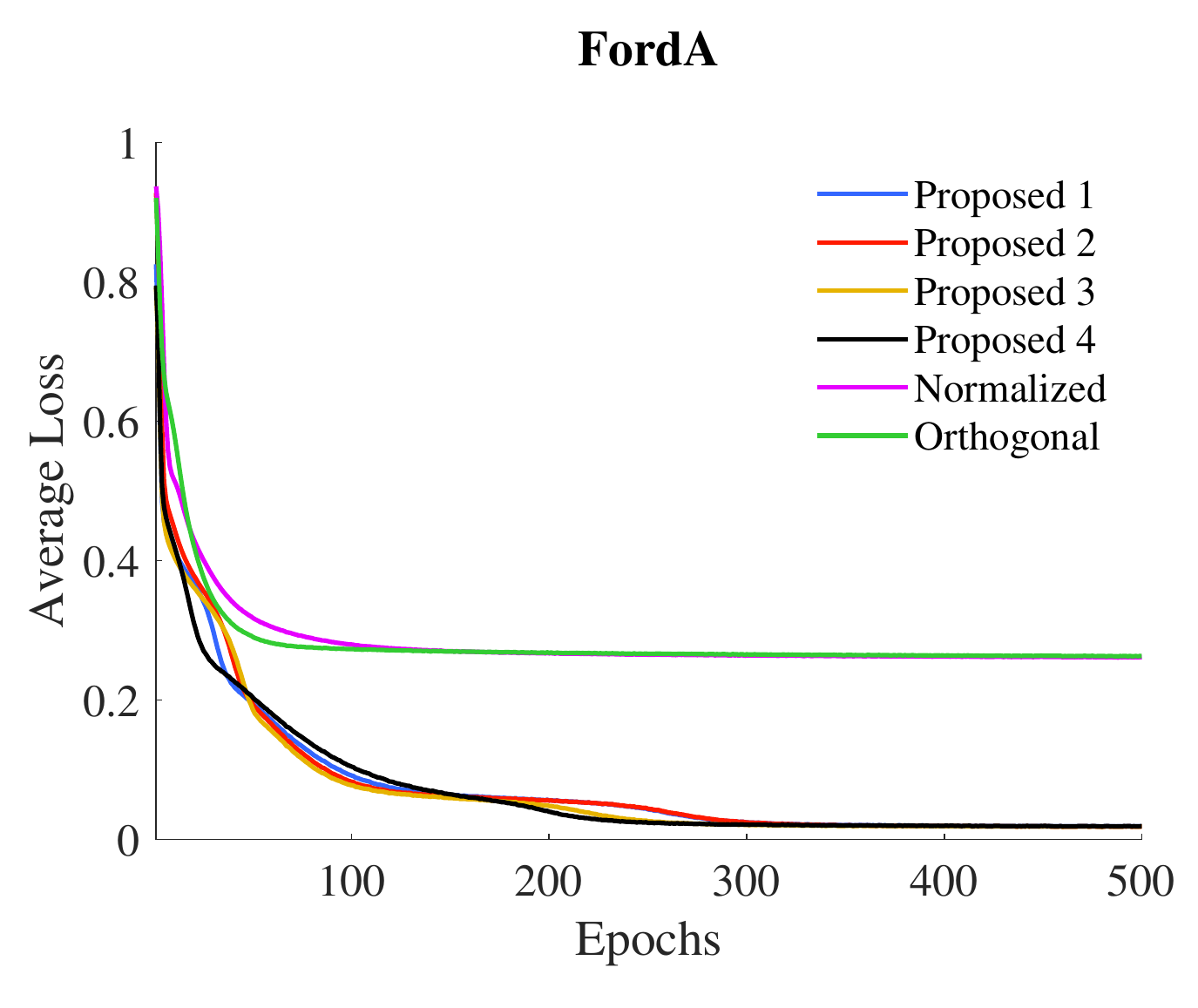}}
\end{subfigure}
\begin{subfigure}[t]{0.47\textwidth}
\raisebox{-\height}{\includegraphics[scale=0.42]{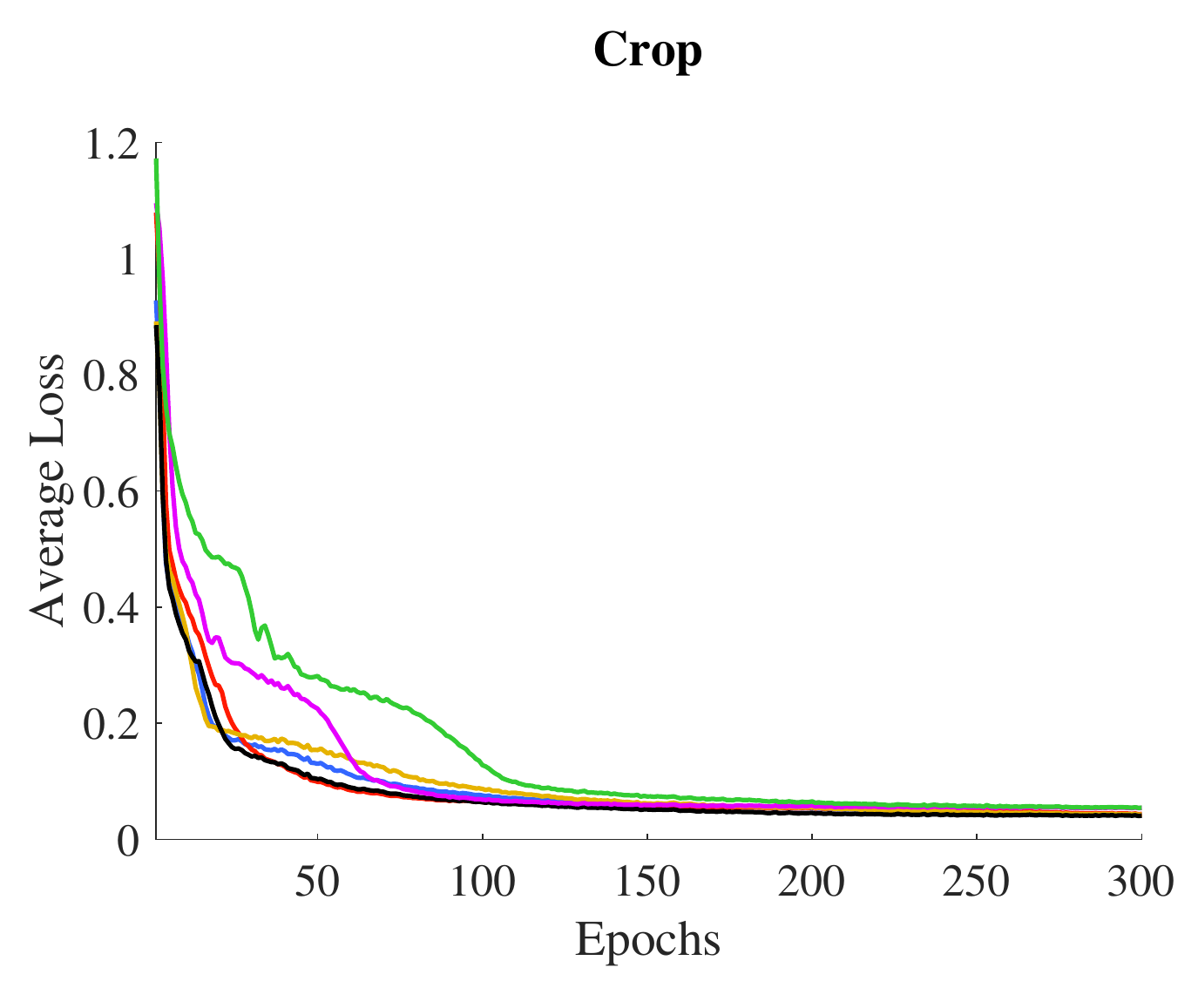}}
\end{subfigure}
\begin{subfigure}[t]{0.47\textwidth}
\raisebox{-\height}{\includegraphics[scale=0.42]{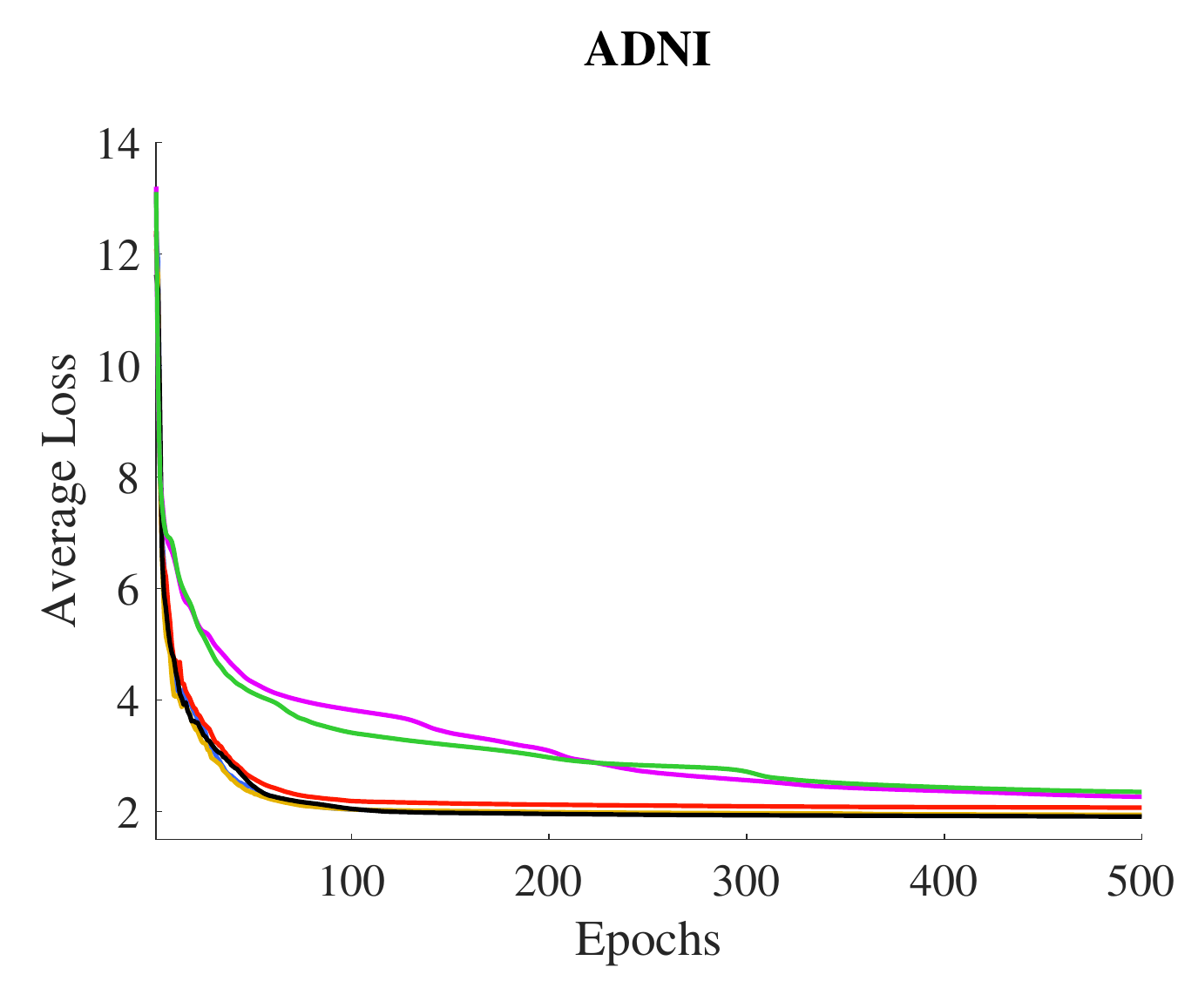}}
\end{subfigure}
\caption{The training loss of the different methods applied to the univariate and multivariate datasets.}
\label{figure1}
\end{figure}

Figure~\ref{figure1} compares the training loss of the proposed and state-of-the-art initialization methods applied to the univariate and multivariate datasets. As can be seen, the proposed method with any configuration outperforms the prevalent initialization techniques in all experiments, either by achieving a lower loss (ElectricDevices and FordA) or by faster convergence to the same loss (Crop and ADNI).

To further investigate the influence of initialization on the performance, we also evaluate the generalization error in the test set. Table~\ref{table1} reports the test mean square error (MSE) in predicting the feature values per dataset for the utilized initialization methods. As it can be deduced, the proposed initialization method with any configuration achieves superior results to the prevalent initialization approaches, which illustrates the generalizability of the proposed method.

\begin{table}[t]
\caption{The generalization error (MSE) in predicting the feature values for the utilized test sets using the different initialization techniques.}
\label{table1}
\centering
\renewcommand{\arraystretch}{1.25}
\begin{tabular}{lcccc}
\toprule
Method & ElectricDevices & FordA & Crop & ADNI \\ \hline
\midrule
Proposed 1 & 0.935 & 0.038 & 0.086 & 0.465 \\
Proposed 2 & 0.942 & 0.037 & 0.088 & 0.468 \\
Proposed 3 & 0.946 & 0.037 & 0.086 & 0.453 \\
Proposed 4 & 0.895 & 0.038 & 0.082 & 0.449 \\
Normalized & 1.012 & 0.523 & 0.109 & 0.473 \\
Orthogonal & 0.998 & 0.526 & 0.110 & 0.471 \\
\bottomrule
\end{tabular}
\end{table}

More interestingly, the fourth configuration of the proposed method in which the recurrent weights receive more variance than the current input weights outperforms all the other methods in almost all of the experiments.

\section{Conclusion}

In this paper, a robust initialization method was proposed for LSTM networks to address training instability and slow convergence. The proposed method was based on scaled random weights initialization aiming to keep the variance of the network input and output signals in the same range subjected to a number of assumptions simplifying the initialization conditions. The proposed method was applied to univariate and multivariate time series regression datasets and outperformed two state-of-the-art initialization methods in all cases.

The obtained conditions can be optimized for eight or eleven unknowns using a traditional LSTM or peephole LSTM, respectively. In this work, different configurations of the variances were inspected to confirm the proposed assumption for initializing the network weights. Moreover, the proposed method can be used for sequence-to-sequence and sequence-to-label learning paradigms by connecting a fully-connected layer with a desired output size to the LSTM network output. It should also be noted that the initialization conditions need to be properly modified in case of using activation functions other than a hyperbolic tangent, identity function, or logistic sigmoid in the gates.

\subsubsection*{Acknowledgments.} This project has received funding from the European Union's Horizon 2020 research and innovation programme under the Marie Sk{\l}odowska-Curie grant agreement No 721820.

\bibliography{my_ref}

\begin{thebibliography}{10}

\bibitem{Hochreiter2001}
Hochreiter, S., Bengio, Y., Frasconi, P., Schmidhuber, J.:
\newblock Gradient flow in recurrent nets: the difficulty of learning long-term
  dependencies.
\newblock In: A Field Guide to Dynamical Recurrent Neural Networks.
\newblock IEEE Press (2001)

\bibitem{Martens2011}
Martens, J., Sutskever, I.:
\newblock Learning recurrent neural networks with {H}essian-free optimization.
\newblock In: Proceedings of the International Conference on Machine Learning.
  (2011)  1033--1040

\bibitem{Trinh2018}
Trinh, T.H., Dai, A.M., Luong, M.T., Le, Q.V.:
\newblock Learning longer-term dependencies in {RNN}s with auxiliary losses.
\newblock CoRR \textbf{abs/1803.00144} (2018)

\bibitem{Le2015}
Le, Q.V., Jaitly, N., Hinton, G.E.:
\newblock A simple way to initialize recurrent networks of rectified linear
  units.
\newblock CoRR \textbf{abs/1504.00941} (2015)

\bibitem{Vorontsov2017}
Vorontsov, E., Trabelsi, C., Kadoury, S., Pal, C.:
\newblock On orthogonality and learning recurrent networks with long term
  dependencies.
\newblock CoRR \textbf{abs/1702.00071} (2017)

\bibitem{Hochreiter1997}
Hochreiter, S., Schmidhuber, J.:
\newblock Long short-term memory.
\newblock Neural Computation \textbf{9}(8) (1997)  1735--1780

\bibitem{Cho2014}
Cho, K., van Merrienboer, B., Gulcehre, C., Bahdanau, D., Bougares, F.,
  Schwenk, H., Bengio, Y.:
\newblock Learning phrase representations using {RNN} encoder{--}decoder for
  statistical machine translation.
\newblock In: Proceedings of the 2014 Conference on Empirical Methods in
  Natural Language Processing. (2014)  1724--1734

\bibitem{Sutskever2014}
Sutskever, I., Vinyals, O., Le, Q.V.:
\newblock Sequence to sequence learning with neural networks.
\newblock In: Advances in neural information processing systems. (2014)
  3104--3112

\bibitem{Gers2002}
Gers, F.A., Schraudolph, N.N., Schmidhuber, J.:
\newblock Learning precise timing with {LSTM} recurrent networks.
\newblock Journal of Machine Learning Research \textbf{3} (2002)  115--143

\bibitem{Dau2018}
Dau, H.A., Bagnall, A., Kamgar, K., Yeh, C.C.M., Zhu, Y., Gharghabi, S.,
  Ratanamahatana, C.A., Keogh, E.:
\newblock The {UCR} {T}ime {S}eries {A}rchive.
\newblock CoRR \textbf{abs/1810.07758} (2018)

\bibitem{Ghazi2019}
Ghazi, M.M., Nielsen, M., Pai, A., Cardoso, M.J., Modat, M., Ourselin, S.,
  S{\o}rensen, L.:
\newblock Training recurrent neural networks robust to incomplete data:
  Application to {A}lzheimer's disease progression modeling.
\newblock Medical Image Analysis \textbf{53} (2019)  39--46

\bibitem{Petersen2010}
Petersen, R.C., Aisen, P.S., Beckett, L.A., Donohue, M.C., Gamst, A.C., Harvey,
  D.J., Jack, C.R., Jagust, W.J., Shaw, L.M., Toga, A.W., Trojanowski, J.Q.,
  Weiner, M.W.:
\newblock Alzheimer's {D}isease {N}euroimaging {I}nitiative ({ADNI}): clinical
  characterization.
\newblock Neurology \textbf{74} (2010)  201--209

\bibitem{Glorot2010}
Glorot, X., Bengio, Y.:
\newblock Understanding the difficulty of training deep feedforward neural
  networks.
\newblock In: Proceedings of the International Conference on Artificial
  Intelligence and Statistics. (2010)  249--256

\bibitem{Erhan2009}
Erhan, D., Manzagol, P.A., Bengio, Y., Bengio, S., Vincent, P.:
\newblock The difficulty of training deep architectures and the effect of
  unsupervised pre-training.
\newblock In: Proceedings of the International Conference on Artificial
  Intelligence and Statistics. (2009)  153--160

\bibitem{He2015}
He, K., Zhang, X., Ren, S., Sun, J.:
\newblock Delving deep into rectifiers: Surpassing human-level performance on
  {ImageNet} classification.
\newblock In: Proceedings of the 2015 IEEE International Conference on Computer
  Vision. (2015)  1026--1034

\bibitem{Talathi2015}
Talathi, S.S., Vartak, A.:
\newblock Improving performance of recurrent neural network with {ReLU}
  nonlinearity.
\newblock CoRR \textbf{abs/1511.03771} (2015)

\bibitem{Buraczewski2016}
Buraczewski, D., Damek, E., Mikosch, T.,  et~al.:
\newblock Stochastic models with power-law tails.
\newblock Springer (2016)

\end{thebibliography}
\bibliographystyle{splncs}

\end{document}